\pdfoutput=1

\documentclass[11pt]{article}

\usepackage[final]{acl}

\newif\ifdraft\drafttrue
\newif\ifanon\anonfalse
\newif\ifincludeappendix\includeappendixtrue

\usepackage{times}
\usepackage{latexsym}

\usepackage[T1]{fontenc}

\usepackage[utf8]{inputenc}

\usepackage{microtype}

\usepackage{inconsolata}

\usepackage{amsmath}
\usepackage{hyperref}       
\usepackage{url}            
\usepackage{booktabs}       
\usepackage{amsfonts}       
\usepackage{nicefrac}       
\usepackage{microtype}      
\usepackage{xspace}
\usepackage{graphicx}
\usepackage{cleveref}
\usepackage{listings}
\usepackage{caption}
\usepackage{float}
\usepackage{subcaption}
\usepackage{paralist}
\usepackage{multirow}
\usepackage{titlesec}
\usepackage{amsfonts} 

\lstdefinestyle{codeblock}{
    basicstyle=\ttfamily\footnotesize,
    frame=lines,
    backgroundcolor=\color{gray!5!white},
    commentstyle=\color{red!60!black},
    keywordstyle=\color{green!50!black},
    stringstyle=\color{red!60!black},
    basicstyle=\ttfamily\footnotesize,
    breakatwhitespace=false,         
    breaklines=true,                 
    captionpos=b,                    
    keepspaces=true,                 
    showspaces=false,                
    showstringspaces=false,
    showtabs=false,                  
    tabsize=2,
    escapechar={~},
}

\definecolor{doctestcolor}{RGB}{230, 239, 255}
\definecolor{errorcolor}{RGB}{255,232,236}
\definecolor{signaturecolor}{RGB}{252, 232, 235}

\Crefname{section}{\S}{\S\S}
\crefformat{section}{\S#2#1#3}
\Crefname{figure}{Figure}{Figures}
\crefformat{figure}{Figure #2#1#3}
\Crefname{Figure}{Figure}{Figures}
\Crefname{Table}{Table}{Tables}
\crefformat{table}{Table #2#1#3}

\title{Creating and Repairing Robot Programs in Open-World Domains}

\author{Claire Schlesinger\ \\
  Northeastern University\thanks{Work primarily done while at an internship at University of Texas at Austin} \\
  \small{schlesinger.e@northeastern.edu} \\\And
  Arjun Guha \\
  Northeastern University \\
  \small{a.guha@northeastern.edu} \\\And
  Joydeep Biswas \\
  University of Texas at Austin \\
  \small{joydeep.b@cs.utexas.edu}}

\newcommand{\RoboRepair}{\textsc{RoboRepair}}
\newcommand{\thebenchmark}{\RoboRepair{} Benchmark}
\usepackage{multirow}

\begin{document}

\maketitle

\begin{abstract}
Using Large Language Models (LLMs) to produce robot programs from natural language has allowed for robot systems that can complete a higher diversity of tasks.  However, LLM-generated programs may be faulty, either due to ambiguity in instructions, misinterpretation of the desired task, or missing  information about the world state. As these programs run, the state of the world changes and they gather new information. When a failure occurs, it is important that they recover from the current world state and avoid repeating steps that they they previously completed successfully. We propose \RoboRepair{}, a system which traces the execution of a program up until error, and then runs an LLM-produced \emph{recovery program} that minimizes repeated actions.

To evaluate the efficacy of our system, we create a benchmark consisting of eleven tasks with various error conditions that require the generation of a recovery program. We compare the efficiency of the recovery program to a plan built with an oracle that has foreknowledge of future errors.
\end{abstract}





\section{Introduction}

Service robots are general-purpose robots that are expected to be able to complete a variety of tasks in a home or office setting, including novel tasks that roboticists did not foresee. Towards this goal, there has been  significant work on using LLMs to turn natural language instructions into short plans or \emph{task programs} to execute on behalf of non-expert end-users~\cite{codebotler, Liu2023LLMPEL, driess2023palmeembodiedmultimodallanguage}. The advantage of LLMs-based approaches is that (1)~they are significantly better than prior work at translating natural language to code, and (2)~they have open-world knowledge that allows them to fill in commonsense details that may be missing from natural language instructions provided by non-experts.

However, these LLM-authored task programs often go wrong for a variety of reasons. (1)~Even the most capable models make errors, or misinterpret users' intentions; (2)~programs may make faulty assumptions about the world state; and (3)~users may need to provide additional instructions or information to help the robot complete the task successfully. The na\"ive solution to these problems is to re-execute the task program, but re-execution may not be possible in every failing scenario, or may be sub-optimal if it involves re-executing completed actions. In contrast, when a human is given a simple task, they are often able to gracefully handle unforeseen situations by taking corrective actions or asking for help themselves. 

\begin{table*}[t]
    \centering
    \begin{tabular}{p{75mm}|p{75mm}}
         Task Description & Failure Mode \\
         \hline
         Gather all the books in the house and return them to where they belong.
         & Misinterpret user intentions: "where they belong" is ambiguous and could mean either an office or a bedroom. \\
         \hline
         Go to every conference room and count how many chairs each of them have & Faulty assumptions about world state: The robot may assume every conference rooms is accessible, but if any are locked, the task would fail. \\
         \hline
         Ask everyone what kind of chips they want and gather enough bags of each for everyone. We have Doritos and Takis. & User could provide additional information: As the robot is in the process of the task, a user could interrupt and inform the robot that they had bought Cheetos and to add that option to the list of chips.
    \end{tabular}
    \caption{Task descriptions and possible failure types associated with that task.}
    \label{tab:failure-descriptions}
\end{table*}

We present \RoboRepair{}, an approach to natural language programming for service robots that supports both \emph{automatic and interactive failure recovery}. Our two key insights are that (1)~a failure in a task program may not manifest immediately, i.e., a failure may not cause an immediate program exception; and (2)~in real-world deployments the number of possible failure modes is enormous, which makes it very challenging for an LLM to generate a program that is robust to all possible errors in a single shot. \RoboRepair{} addresses both of these problems by first prompting the LLM to generate simple task programs that assume no errors occur. Under the hood, the runtime system traces program execution and interrupts the programs when it detects an error, or when the user interrupts the program themselves. To resume execution, \RoboRepair{} synthesizes a \emph{recovery program} that resumes execution from a context that includes the prior execution trace and feedback about the error. To avoid needlessly re-executing prior actions, \RoboRepair{} adds the execution trace to the prompt, to condition generation, and reifies it as runtime state to make it available during re-execution. 

We evaluate \RoboRepair{} with a new benchmark, \thebenchmark{}, which contains eleven problems that require either error handling or creation of recovery programs to solve. We demonstrate that \RoboRepair{} is able to produce recovery programs a majority of the time that perform nearly as well as an optimal plan made by an agent if it had access to an oracle to predict future errors.

\section{Background and Related Work}

\paragraph{LLMs for Robot Tasks}

LLMs make it possible to translate task requests in natural language to actionable steps either as plans, or robot-executable programs. This makes it possible for end-users without programming and robotics expertise to direct a robot to perform novel tasks. Moreover, these natural language directions can be succinct and rely on the LLM's world knowledge to fill in missing details. 

For example, if a household robot is told  \emph{go find my coffee mug}, an LLM can reason that coffee is most likely in the kitchen, and thus should check there first. There are several different approaches for turning this kind of natural language command into actions for a robot, including generating code~\cite{codebotler, singh2022progprompt, 2023liangcodeaspolicies, wang2023demo2codesummarizingdemonstrationssynthesizing}, generating plans~\cite{driess2023palmeembodiedmultimodallanguage, irpan2022doasIsay, song2023llmplannerfewshotgroundedplanning, huang2022inner}, and generating specifications~\cite{liu2023lang2ltl, Liu2023LLMPEL, Maoz2015GR1LTL}. LLM+P uses LLMs to write PDDL \cite{McDermott1998PDDLthePD} specifications to solve human given tasks in the most optimal manner rather than relying on the LLM to generate the plan.

\RoboRepair{} builds on the CodeBotler~ \citep{codebotler} platform which uses an LLM to produce task programs from natural language instructions. CodeBotler prefixes the prompt with a description of the robot API and a few shot examples. In contrast,  \RoboRepair{} augments the prompt with prior state and few-shot examples of generated task programs, their traces (including errors), and recovery programs. \RoboRepair{} generates both the initial task program and recovery program for any natural language task description.

\paragraph{Handling Changes to the Task and Environment} The aforementioned approaches do not gracefully adapt when the LLM-generated program or plan goes wrong. For example, consider the world state where the user's coffee mug is on the dining table: a program/plan that only looks in the kitchen will fail, and the goal of \RoboRepair{} is to overcome such failures, which may occur by misinterpreting of the task, making faulty assumptions about the world state, or missing information from users. \Cref{tab:failure-descriptions} shows an example of some tasks and possible failure modes.

There is some work on using LLMs to create self-correcting plans. Inner Monologue~\cite{huang2022inner} plans one action at a time and provides feedback after executing every action. However, Inner Monologue does not account for spontaneous human feedback such as modification to the original task. PaLM-E~\citep{driess2023palmeembodiedmultimodallanguage} also produces a plan one action at a time, but is sensitive to environmental changes. PaLM-E can change plans, including repeating steps, if the environment spontaneously changes. LLM-Planner~\citep{song2023llmplannerfewshotgroundedplanning} produces a high level plan in one shot, and as its executed, tracks what has been completed and updates from the environment to produce new plans if the robot is unable to complete an action. RTPL~\citep{hammond2019automaticfailurerecoveryenduser} utilizes a Bayesian network to track possible errors and if an error occurs, uses that network to identify the cause of the error and re-execute the right portion of the program for recovery. While all these works attempt to resume recovery from the broken state and circumvent either robot errors, environmental errors, or the need for human feedback, none do all three, and none are able to take spontaneous human feedback.

\paragraph{LLMs for Automated Program Repair}

LLMs have been used for \emph{automated program repair} in software engineering~\citep{zhang2024criticalreviewlargelanguage, sun2024llmruntimeerrorhandler, xia2022zeroshotapr, xia2023aprllmpretrain, xia2023conversationgoingfixing162}.  ChatRepair~\citep{xia2023conversationgoingfixing162} uses ChatGPT to repair buggyg programs, using compiler and test errors as feedback. \RoboRepair{} is most closely related to Healer~\citep{zhang2024criticalreviewlargelanguage}, which uses an LLM to generate new code to respond to runtime errors instead of patching existing code.

There are many APR benchmarks for general software engineering tasks, e.g., DebugBench~\citep{tian2024debugbenchevaluatingdebuggingcapability} is based on LeetCode solutions and QuixBugs~\citep{lin2017quixbugs} has hand-crafted challenge problems. Our benchmark is designed for robots tasks and runs in a simulated environment, using execution traces to determine task success, similar to RoboEval~\citep{codebotler}.

\section{\RoboRepair{}}

\begin{figure*}[t]
    \centering
    \includegraphics[width=\linewidth]{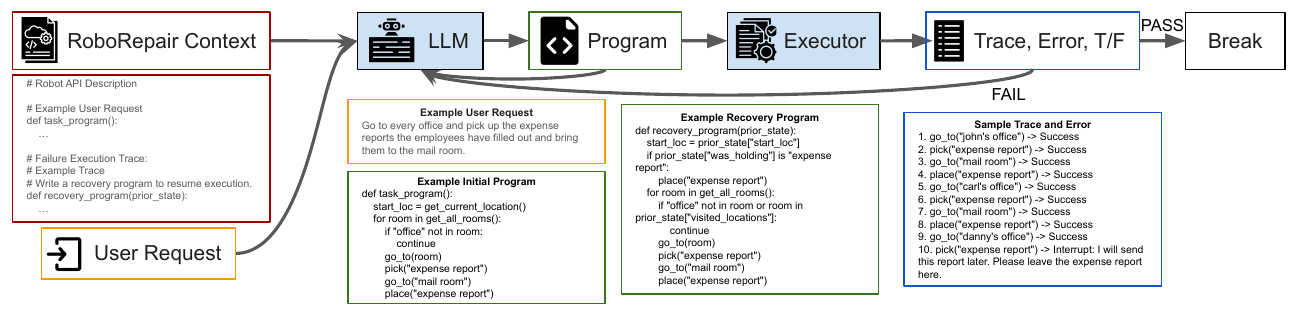}
    \caption{Diagram and Example of \RoboRepair{}}
    \label{fig:roborepair}
\end{figure*}

In the simplest case, \RoboRepair{} turns a user's natural language request into a single program that runs on the robot and completes the task. However, \RoboRepair{} is designed to gracefully recover from errors and interruptions that may occur during program execution. Thus given a natural language request, \RoboRepair{} generates and executes a sequence of robot programs, where each subsequent program is intended to recover from the failure encountered by the previous program. To generate each program, \RoboRepair{} prompts an LLM with 1)~the original task, and the interleaved sequence of 2)~previously executed programs, 3)~their execution traces, and 4)~errors or new requests from users they encountered.~\Cref{fig:roborepair} visualizes the process \RoboRepair{} goes through to generate the initial and recovery program.

\begin{figure*}[t]
    \begin{subfigure}[b]{\textwidth}
        \centering
        \includegraphics[width=\linewidth]{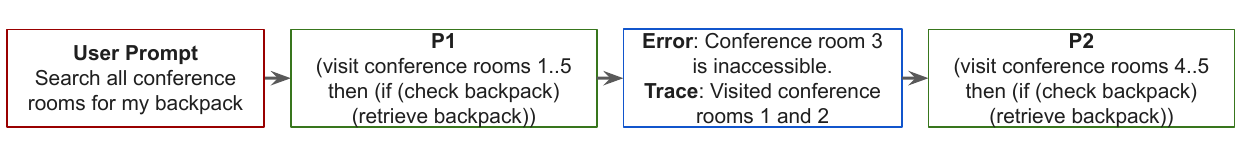}
        \caption{\RoboRepair{} recovering a task that failed due to an error triggered by the robot API.}
        \label{fig:roborepair-ex-1}
    \end{subfigure}
    \begin{subfigure}[b]{\textwidth}
        \centering
        \includegraphics[width=\linewidth]{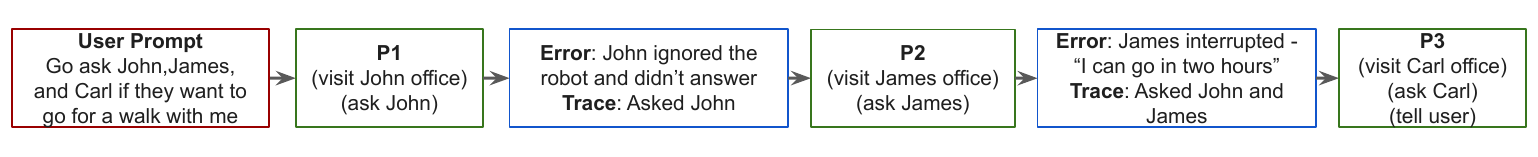}
        \caption{\RoboRepair{} recovering a task that failed due to an error triggered by the robot API and an error caused by a human interrupt}
        \label{fig:roborepair-ex-2}
    \end{subfigure}
    \caption{Examples of \RoboRepair{} recovering from errors.}
    \label{fig:roborepair-ex}
\end{figure*}

For example, \cref{fig:roborepair-ex-1} showcases how \RoboRepair{} handles an error triggered by the robot API. The robot is asked to visit every conference room to find and retrieve a backpack. After checking rooms 1 and 2, the robot finds that room 3 is inaccessible. Instead of failing the task, the robot recovers and completes the task by checking the remaining conference rooms. Notably, it does not re-check rooms 1 and 2 because the trace indicates that they were already visited before recovery. In \cref{fig:roborepair-ex-2}, the robot is requested to see if John, James, or Carl wants to go for a walk. However, John does not answer the question (causing a time out), and James interrupts to provide specific details for his response. \RoboRepair{} uses the information about the errors/feedback from John and James to update its final response. It assumes John can't go on a walk because he couldn't answer and will pass on the information provided by James in the final report to the original user. 

\subsection{Traces and Execution}

A \emph{trace} of a robot program is a tuple containing 1)~a list of all locations visited by the robot, 2)~the object the robot is currently carrying, 3)~the questions the robot asked, and 4)~the list of all objects detected in each visited location.
\begin{equation*}
    r_\pi = \langle [loc_{0:t}], obj, [\langle q, a \rangle_{1:t}], [\langle loc, obj \rangle_{1:t}]\rangle
\end{equation*}

\begin{figure}[t]
    \centering
    \begin{lstlisting}[style=codeblock]
1. go_to("john's office") -> Success
2. pick("expense report") -> Success
3. go_to("mail room") -> Success
4. place("expense report") -> Success
5. go_to("danny's office") -> Success
6. pick("expense report") -> Interrupt: 
I will send this report later.\end{lstlisting}
    \caption{The natural language form of the trace for \cref{fig:example-task-program}. The trace highlights the steps the robot took and includes the message provided by the Danny when he interrupted the task.}
    \label{fig:example-nl-trace}
\end{figure}

We represent traces in two forms:~1) in a natural language format which lists it as a series of steps (e.g., \cref{fig:example-nl-trace}) and 2)~as an object available at runtime to the robot program. The two representations allows for the language model to both understand the steps that occurred and use code to accurately recreate any internal state that is necessary after recovery.

As a robot program, $\pi \in \Pi$, is executed by an executor \texttt{Exec} on a world $w \in W$, it produces a trace, $r \in R$, an error, $e \in E$, and \texttt{PASS} if the program executed successfully and \texttt{FAIL} otherwise.
\begin{equation*}
    \mathrm{Exec}: W \times \Pi \rightarrow W \times R \times E \times \{\texttt{PASS}, \texttt{FAIL}\}
\end{equation*}


\texttt{Exec} also produces an error message. This error message can either be from a robot API call failing or it can be given by a human if they interrupt the task. If a human interrupts the task, they can provide additional instructions or corrections to issues that could not have been previously known.


\subsection{Generating Robot Programs}


\RoboRepair{} produces a new program by utilizing a task prompt, $t \in T$, the sequence of all previously generated programs for that prompt, the sequence of traces produced by those programs, and the sequence of errors produced by those programs:
\begin{eqnarray*}
\RoboRepair{} :  T \times R^* \times E^* \times \Pi^* \rightarrow \Pi \\
    \RoboRepair{}(p, r_{[0:n-1]}, e_{[0:n-1]}, \pi_{[0:n-1]}) = \pi_n
\end{eqnarray*}

\RoboRepair{} operates recursively where every produced program and its result can be used to produce the next program in the sequence. To generate the original program, we define the base case with an empty trace, error message, and previous program, 
$r_0  = \langle \rangle, e_0  = \verb|""|, \pi_0 =  \verb|""|.$

\subsubsection{Context and Prompting}

\RoboRepair{} uses an LLM to generate robot programs. It uses a context which provides examples consisting of an original natural language prompt, a task program, its resulting trace and error, and the expected recovery program. The context also contains the robot API definition (\cref{fig:robot-api}) along with the definition of the trace object (\cref{fig:prior-state-def}). 

\begin{figure}[t]
    \centering
    \begin{lstlisting}[style=codeblock]
# Get the current location of the robot.
def get_current_location() -> str:
# Get a list of all rooms.
def get_all_rooms() -> list[str]:
# Check for an object.
def is_in_room(object : str) -> bool:
# Go to a specific named room, 
def go_to(room : str) -> None:
# Ask a person a question.
def ask(person : str, question : str, options: list[str]) -> str:
# Say the message out loud.
def say(message : str) -> None:
# Pick up an object if none are held.
def pick(obj: str) -> None:
# Place an object down if held.
def place(obj: str) -> None:\end{lstlisting}
    \caption{Robot API definition provided in the \RoboRepair{} context.}
    \label{fig:robot-api}
\end{figure}

\begin{figure}[t]
    \centering
    \begin{lstlisting}[style=codeblock]
prior_st = {
  # A list of all visited locations
  "visited_locations": List[str], 
  # The starting location
  "start_loc": str, 
  # Responses to questions
  "responses": List[Dict[str, obj]], 
  # Where every object was seen.
  "seen": Dict[str, List[str]], 
  # What the robot is holding
  "was_holding": Union[str, None] 
}\end{lstlisting}
    \caption{Prior state definition provided in the \RoboRepair{} context.}
    \label{fig:prior-state-def}
\end{figure}

\subsubsection{Task Program Generation}

In the base case, where \RoboRepair{} is only provided with a natural language task description, \RoboRepair{} generates the initial robot program that may complete that task. To do this, the prompt is formatted by providing the natural language direction as a comment, and then adding the \texttt{task\_program} function header (\cref{fig:task-program-prompt}). The prompt is appended to the end of the \RoboRepair{} context and then we prompt the LLM to generate the completion.

\begin{figure}[t]
    \centering
    \begin{lstlisting}[style=codeblock]
# Go to every office and pick up the 
# expense reports the employees have 
# filled out and bring them to the 
# mail room.
def task_program():\end{lstlisting}
    \caption{Example \texttt{task\_program} prompt for the natural language input: "Go to every office and pick up the expense reports the employees have filled out and bring them to the mail room".}
    \label{fig:task-program-prompt}
\end{figure}

\RoboRepair{} generates an initial program $\pi$ to complete the task (\cref{fig:example-task-program}). The prompt may not highlight all the steps needed for the task nor all the information about the world. The generated program should utilize the robot API and use the LLM's open world knowledge to fill in any missing details. The program should be made so that upon execution, \texttt{Exec} returns \texttt{PASS}. The initially generated program is based on the LLM's best guess of the world state and may run into errors that have to be processed later.

\begin{figure}[t]
    \centering
    \begin{lstlisting}[style=codeblock]
def task_program():
  start_loc = get_current_location()
  for room in get_all_rooms():
    if "office" not in room:
      continue
    go_to(room)
    pick("expense report")
    go_to("mail room")
    place("expense report")\end{lstlisting}
    \caption{Example task program for the prompt ``Go to every office and pick up the expense reports the employees have filled out and bring them to the mail room.''}
    \label{fig:example-task-program}
\end{figure}

\subsubsection{Recovery Program Generation}

When an error occurs, either because the generated code is faulty,  the environment does not behave as assumed, or the user interrupts the program, \RoboRepair{} generates a new program (\cref{fig:recovery-program-example}), $\pi'$ that tries to avoid repeating steps that were successfully completed in the previous program $\pi$.

To generate this program, the original prompt is extended by adding the sequence of already produced programs, their traces, and their errors.~\cref{fig:recovery-prompt} shows what the prompt for a recovery program would look like for the first recovery program is generated. This prompt is appended to the end of the \RoboRepair{} context.

\begin{figure}[t]
    \centering
    \begin{lstlisting}[style=codeblock]
# Go to the office and ask my husband if
# he is ready for dinner.
def task_program():
    ...
# The program had the following trace:
# 1. go_to("office") -> Success
# 2. ask("Are you ready for dinner?",
#    ["yes", "no"]) -> Error: timed out.
# Generate recovery program
def recovery_program(prior_st):\end{lstlisting}
    \caption{Example of a prompt when generating a recovery program.}
    \label{fig:recovery-prompt}
\end{figure}

If we had perfect foreknowledge of the world state, including future errors, an optimal plan $\Gamma$ could be formed such that it completes all required steps of the task. \RoboRepair{} should produce $\pi'$ such that  when $\pi'$ is executed, the difference between the sum of the lengths of the traces from $\pi$ and $\pi'$, and the length of $\Gamma$ is minimized, i.e., 
$\pi' = \arg_{\pi'}\min \left(|r_\pi| + |r_{\pi'}| - |\Gamma|\right)$.

\begin{figure}[t]
    \centering
    \begin{lstlisting}[style=codeblock]
def recovery_program(prior_st):
  start_loc = prior_st["start_loc"]
  if prior_st["was_holding"] \
    is "expense report":
    place("expense report")
  for room in get_all_rooms():
    if "office" not in room \
        or room in \
        prior_st["visited_locations"]:
      continue
    go_to(room)
    pick("expense report")
    go_to("mail room")
    place("expense report")\end{lstlisting}
    \caption{Example of a recovery program produced for the task shown in~\cref{fig:example-task-program} produced by the trace in~\cref{fig:example-nl-trace}. The recovery program skips already visited rooms and returns the document if it had successfully picked it up before being interrupted.}
    \label{fig:recovery-program-example}
\end{figure}

\subsection{Error Handling}

Different types of errors carry different kinds of information. When an error occurs because a user interrupts execution, \RoboRepair{} must regenerate the program to fit the user's feedback. For well-defined API errors, it is possible for \RoboRepair{} to produce code that automatically handles errors instead of generating a new program.

To enable the latter kind of error handling, we label the robot API with the errors that each action could throw, and include error handling code in the few-shot examples of programs. This allows \RoboRepair{} to avoid generating recovery programs for every kind of error.

\begin{table*}[!t]
    \centering
    \footnotesize
    \begin{tabular}{l|p{55mm}|p{55mm}}
        \textbf{Problem} & \textbf{Description} & \textbf{Failure Mode} \\
        \hline
         FindBackpack & Search through all the conference rooms to find a backpack. & One of the conference rooms will be inaccessible. \\
         \hline
         HalloweenShopping & Visit every office and asking everyone how many chocolates, gummies, and caramels they want. & One of the people in the office will refuse to answer any questions. \\
         \hline
         WalkingBuddy & Ask a series of people if they want to go for a walk. & The first person will not answer the question and the second person will not be accessible. \\
         \hline
         SearchHarder & Look through every room in the house and find a book. & The robot is unable to find the book, a human interrupts it and tells to check the bedroom again. \\
         \hline
         BringCupsInterrupt & Go to every room and bring all cups to the kitchen. & A human will interrupt and request to keep their cup. \\
         \hline
         MeetingTime & Ask John whether to meet at 6 or 7. & John will interrupt to say he can only meet at 3 \\
         \hline
         PlacePlants & Take the plants and place them into every office. & A person will interrupt to say they don't want the plant. \\
         \hline
         BadWeather & Tell everyone in the living room and game room to go outside. & A human will interrupt the robot and tell it to also mention that it will rain later. \\
         \hline
         GoToBanned & Check every warehouse for a box of chocolate and report back. & The robot will be interrupted when attempting to enter warehouse 4 and told to not enter warehouses 4-6. \\
         \hline
         WhichPopcorn & Go to every dorm and ask them if they want cheese or plain popcorn. & A human will interrupt the robot to tell it to include kettle corn in its options. \\
         \hline
         BringCoffee & Bring me a cup of coffee. & The coffee will be stored in the office rather than the kitchen, so if the robot fails to find it, the human will interrupt to provide that information.
    \end{tabular}
    \caption{The problems in the \thebenchmark{}}
    \label{tab:benchmark-problems}
\end{table*}

\section{Evaluation}

We empirically seek to answer three questions to  evaluate \RoboRepair{}:
1) How often does \RoboRepair{} successfully complete a task after a recovery?
2) What is the optimality of the execution traces generated by running \RoboRepair{}-generated programs?
3) How does the choice of different language models and inference modes affect success rates and optimality of \RoboRepair{}-generated programs?
We first present the evaluation benchmark that we use to answer the above questions. For the first two evaluation questions, we use GPT-4o \cite{openai2024gpt4technicalreport} as the LLM component which generates task programs. We generate code using top-p sampling (0.95), sampling with temperature (0.2), and with a limited number of output tokens (512). For each prompt we generate 20 completions.

\RoboRepair{}'s prompt contains information about the kind of errors the robot API can throw and how to write proper error handling. This enables \RoboRepair{} to circumvent the need for program regeneration saving on the cost of running a LLM. However, to see how this effects the performance, we run an ablation of \RoboRepair{} where the context lacks information about error handling.

\renewcommand{\floatpagefraction}{0.99}  
\renewcommand{\textfraction}{0.1}       
\renewcommand{\topfraction}{0.99}        

\subsection{The \thebenchmark{}}

To evaluate the efficacy of \RoboRepair{} we developed the \thebenchmark{} consisting of eleven tasks each with five unique prompts and an exception that requires construction of a recovery program. The descriptions of these tasks can be found in~\cref{tab:benchmark-problems}. The exceptions range from user interrupts to errors that might occur in execution of the robot API. Each task is accompanied by constraints that must be fulfilled for it to be considered completed. Three tasks have errors triggered by the robot API and eight have user-given interrupts.

Each benchmark task has a set of constraints $c \in C$ which must all be satisfied for the task to be considered successfully completed. We construct an evaluator \texttt{Eval} which is able to take the collection of traces for all generated programs and determine if the sum total fits the constraints. These constraints test that certain actions occurred at the right time and that critical actions only occurred once.

\begin{equation*}
    \texttt{Eval}: R^* \times C \rightarrow \{\texttt{SAT}, \texttt{UNSAT}\}
\end{equation*}

\begin{figure}[t]
    \centering
    \begin{lstlisting}[style=codeblock]
def recovery_program(prior_st):
  start_loc = prior_st["start_loc"]
  if prior_st["was_holding"] == "plant":
    go_to(start_loc)
    place("plant")
  for room in get_all_rooms():
    if "office" in room and \
        room not in \
        prior_st["visited_locations"]:
      pick("plant")
      go_to(room)
      if is_in_room("person"):
        say("I have a plant for you.")
        response = ask("person", 
            "Keep Plant?", 
            ["Yes", "No"])
        if response == "No":
          go_to(start_loc)
          place("plant")
          continue
      place("plant")
      go_to(start_loc)\end{lstlisting}
    \caption{The recovery program after an interruption trying to place a plant in Zarko's office. This response is not optimal because it adds an extra component where it will check for someone and ask them if they want a plant, but it does skip placing plants in offices it previously went to and relocates the plant from Zarko's office.}
    \label{fig:PlacePlants-recovery-example}
\end{figure}

\begin{table*}[t]
\footnotesize
    \centering
    \begin{tabular}{l|r|rr|rr|rr}
        Model & GPT-4o & \multicolumn{2}{|c}{Llama3.1-70B} & \multicolumn{2}{|c}{Llama3.1-8B} & \multicolumn{2}{|c}{DeepSeek-Coder-V2} \\
        \hline
        Inference Mode & Chat & Completion & Chat & Completion & Chat & Completion & Chat \\
        \hline 
        CodeBotler & 16.27\% & 7.36\% & 14.45\% & 7.64\% & 4.73\% & 9.25\% & 4.27\% \\
        Ablation & 43.09\% & 26.82\% & 31.27\% & 25.45\% & 21.45\% & \textbf{14.45\%} & 12.36\% \\
        \RoboRepair{} & \textbf{57.18\%} & 30.91\% & \textbf{42.47\%} & \textbf{25.73\%} & 21.36\% & \textbf{14.45\%} & 11.91\% \\
    \end{tabular}
    \caption{Pass@1 rates for GPT-4o, Llama3.1-70B, Llama3.1-8B, and DeepSeek-Coder-V2-Lite with \RoboRepair{}, without error handling, and with error handling added to CodeBotler.}
    \label{tab:pass@1-roborepair}
\end{table*}

\begin{table*}
\footnotesize
    \centering
    \begin{tabular}{lrr|rr|rr|rr}
        &&& \multicolumn{2}{|c}{\RoboRepair{}} & \multicolumn{2}{|c}{Ablation} & \multicolumn{2}{|c}{Error Handling} \\
        \hline
        & Optimal & Perfect & Avg. & & Avg. & & Avg. &  \\
        Problem & Steps & Steps & Steps & \% Diff. & Steps & \% Diff. & Steps & \% Diff. \\
        \hline 
        BadWeather & 6 & 6 & 7.69 & 28.14\% & \textbf{7.11} & \textbf{18.50\%} & -- & -- \\
        BringCoffee & 4 & 4 & -- & -- & -- & -- & -- & -- \\
        BringCupsInterrupt & 10 & 14 & \textbf{20.00} & \textbf{100.00\%} & -- & -- & -- & -- \\
        FindBackpack & 7 & 9 & 9.09 & 29.86\% & \textbf{8.92} & \textbf{27.43\%} & 9.00 & 28.57\% \\
        GoToBanned & 14 & 20 & \textbf{15.00} & \textbf{7.14\%} & \textbf{15.00} & \textbf{7.14\%} & -- & -- \\
        HalloweenShopping & 12 & 18 & 15.98 & 33.20\% & 18.89 & 57.41\% & \textbf{15.91} & \textbf{32.59\%} \\
        MeetingTime & 5 & 5 & -- & -- & -- & -- & -- & -- \\
        PlacePlants & 11 & 15 & \textbf{17.65} & \textbf{60.45\%} & -- & -- & -- & -- \\
        SearchHarder & 14 & 9 & \textbf{14.35} & \textbf{2.48\%} & 15.11 & 7.90\% & -- & -- \\
        WalkingBuddy & 5 & 11 & \textbf{9.00} & \textbf{80.00\%} & -- & -- & -- & -- \\
        WhichPopcorn & 20 & 20 & \textbf{20.00} & \textbf{0.00\%} & \textbf{20.00} & \textbf{0.00\%} & -- & -- \\
    \end{tabular}
    \caption{The difference between the average steps taken to complete a task versus the optimal steps possible with perfect knowledge for both \RoboRepair{}, CodeBotler with the error handling context and our \RoboRepair{} ablation. ``--'' means that the task did not complete. ``Perfect steps'' assumes no interruptions or errors.}
    \label{tab:optimal-results}
\end{table*}

\begin{table*}[!t]
\footnotesize
    \centering
    \begin{tabular}{lrr|rrr|rrr}
        &&& \multicolumn{6}{|c}{Llama3.1-70B}  \\
        \hline
        &&& \multicolumn{3}{|c}{Completion} & \multicolumn{3}{|c}{Chat}  \\
        \hline
                & Optimal & Ideal & & Average & & & Average & \\
        Problem & Steps & Steps & Pass@1 & Steps & \% Diff. & Pass@1 & Steps & \% Diff. \\
        \hline 
        BadWeather & 6 & 6 & 0.89 & 7.08 & 17.98\% & \textbf{1.0} & \textbf{7.00} & \textbf{16.67\%} \\
        BringCoffee & 4 & 4 & 0.0 & -- & --\% & 0.0 & -- & --\% \\
        BringCupsInterrupt & 10 & 14 & 0.0 & -- & & 0.0 & -- & --\% \\
        FindBackpack & 7 & 9 & \textbf{1.0} & \textbf{8.52} & \textbf{21.71\%} & \textbf{1.0} & 8.84 & 26.29\% \\
        GoToBanned & 14 & 20 & 0.0 & -- & --\% & 0.0 & -- & --\% \\
        HalloweenShopping & 12 & 18 & 0.0 & -- & --\% & \textbf{0.31} & \textbf{15.77} & \textbf{31.45\%} \\
        MeetingTime & 5 & 5 & 0.0 & -- & --\% & 0.0 & -- & --\% \\
        PlacePlants & 11 & 15 & 0.0 & -- & --\% & 0.0 & -- & --\% \\
        SearchHarder & 14 & 9 & 0.02 & \textbf{15.00} & \textbf{7.14\%} & \textbf{0.52} & 15.33 & 9.48\% \\
        WalkingBuddy & 5 & 11 & 0.49 & \textbf{9.00} & \textbf{80.00\%} & \textbf{0.82} & \textbf{9.00} & \textbf{80.00\%} \\
        WhichPopcorn & 20 & 20 & \textbf{1.0} & \textbf{20.00} & \textbf{0.00\%} & \textbf{1.0} & \textbf{20.00} & \textbf{0.00\%} \\
    \end{tabular}
    \caption{The difference between the average steps taken to complete a task versus the optimal steps possible with perfect knowledge for \RoboRepair{} on Llama3.1-70B in both completion and chat mode. ``--'' means that the task did not complete. ``Perfect steps'' assumes no interruptions or errors.}
    \label{tab:optimal-results-diff-models}
\end{table*}

We compute the pass@1 score \cite{humaneval} across each task over every prompt and generation (\cref{tab:pass@1-roborepair}). We compare against a modified instance of CodeBotler \cite{codebotler} where the context has been changed to note the addition of possible errors in the robot API and to provide few-shot examples with error-handling. \RoboRepair{} outperforms the modified CodeBotler method because it is able to handle user feedback. If CodeBotler continued to run after the user interrupt, it wouldn't be able to change its code to adapt to the new instructions leading to it failing to meet all the constraints for a task. Because of the user-given interrupts, it is necessary to regenerate code for certain tasks as that would be the only way to incorporate the error feedback.

\subsection{Optimality of Recovery}

In this section we compare \RoboRepair{} to plans created using an oracle that has perfect knowledge about the state of the world and exceptions that will occur. This oracle allows us to devise an optimal plan ($\Gamma$). We use Fast Downward~\cite{fastdownward} to construct optimal plans and specify the task and oracle in PDDL~\cite{McDermott1998PDDLthePD}.

\Cref{tab:optimal-results} shows that \RoboRepair{} takes more steps than the optimal plan for most benchmark tasks, which is to be expected. This happens for two reasons: 1)~the optimal plan avoids steps that lead to exceptions, and 2)~when \RoboRepair{} recovers from errors, it takes extra steps to do so.
For example, in the PlacePlants problem, when a human interrupts the task to indicate they don't want the plant, \RoboRepair{} can produce a recovery program that asks every subsequent human if they want the plant (\cref{fig:PlacePlants-recovery-example}). \RoboRepair{} and Codebotler both avoid repeating unnecessary steps, but \RoboRepair{} takes more steps because of its tendency to add extra checks on recovery.

\RoboRepair{} struggles with the MeetingTime task because it is unable to understand the updated instructions included a message for the original user and not the human that was providing updated instructions. \RoboRepair{} also struggles with BringCoffee as it required lots of open-ended knowledge. It tries to go to a room called \texttt{``coffee machine''} which does not exist and produces recovery programs that also have the fault.

\subsection{Variations Between Language Models}

Finally, we compare the performance of \RoboRepair{} across different models. We evaluate Meta Llama 3.1 (8B and 70B)~\cite{dubey2024llama3herdmodels} and DeepSeek-Coder-V2 (16B)~\cite{deepseekai2024deepseekcoderv2breakingbarrierclosedsource} (see \cref{models-and-compute} for licenses and resources used). We evaluated both the base and instruction-tuned versions of these models. (GPT-4o is only available as an instruction-tuned model.) As above, we evaluate both their success rate (pass@1) on tasks, as well as their optimality by comparing their traces to $\Gamma$.

\Cref{tab:pass@1-roborepair} shows that both Llama3.1 and DeepSeek-Coder-V2 perform worse than GPT-4o. Llama3.1-70B benefits from the usage of the chat format while DeepSeek-Coder-V2 and Llama3.1-8B suffer. This is likely due to the larger size of Llama3.1-70B where extra training can improve performance rather than over fitting.

\Cref{tab:optimal-results-diff-models} shows that there is not a significant difference between the optimality of the code generated by the completion models versus code generated by chat models. Both use approximately the same number of steps. What is interesting to note is that Llama 3.1-70B is able to solve an additional task, HalloweenShopping, and is able to produce more successful programs. Similar tables for Llama3.1-8B and DeepSeekCoder-V2-Lite are in \cref{appendix-results}.

\section{Conclusion}

We present \RoboRepair{}, a system which is able to use an LLM to generate a robot program using a robot API to complete a user-provided task, and if that program runs into an error, regenerate a new program which avoids the error. We also created the \thebenchmark{} as a tool for evaluating the performance of \RoboRepair{} on recovering from failures. We tested \RoboRepair{}'s ability to solve the \thebenchmark{} tasks and how optimally it is able to complete those tasks using GPT-4o as the LLM. \RoboRepair{} was able to complete a grand majority of the tasks without repeating steps, but was not the most optimal because the recovery programs often took a more cautious approach to avoid errors going forward. We also tested other LLMs and showed that inference mode can have an effect on the performance of \RoboRepair{}.

\section{Limitations}

This work has several limitations that require further research. The \thebenchmark{} only tested situations where there was a single error and continuation of the task was straightforward. In real world situations a task may encounter several errors as a robot attempts to complete it or there may be tasks where there is a wide range of correct recovery options that are not easily tested. A future benchmark that allows for more errors per task and a higher diversity of correct solutions would be better to test the effectiveness of recovery systems. \RoboRepair{} only recovers in situations where the program receives an error either from the robot API or from a user, but if the robot is performing an action that does not produce an error and a user is unable to stop it, it could enter a state where it is acting erroneously and does not know it needs to repair. Additional research could consider finding ways of using embodied systems to automatically provide feedback that a user might provide if it saw the robot behaving in an undesired fashion.

There are a few risks when automatically generating code including potentially generating hazardous code that can cause damage to the system, or in this case, have the robot perform an action dangerous to a human. 

\section*{Acknowledgements}

This work is partially supported by JP Morgan. Any opinions, findings, and conclusions expressed in this material are those of the authors and do not necessarily reflect the views of the sponsors.

\bibliography{root}

\clearpage

\appendix

\renewcommand{\floatpagefraction}{0.99}  
\renewcommand{\textfraction}{0.01}       
\renewcommand{\topfraction}{0.99} 

\onecolumn

\begin{table*}[t!]
    \centering
    \footnotesize
    \begin{tabular}{lrr|rrr|rrr}
        &&& \multicolumn{6}{|c}{Llama3.1-8B}  \\
        \hline
        &&& \multicolumn{3}{|c}{Completion} & \multicolumn{3}{|c}{Chat}  \\
        \hline
                & Optimal & Ideal & & Average & & & Average & \\
        Problem & Steps & Steps & Pass@1 & Steps & \% Diff. & Pass@1 & Steps & \% Diff. \\
        \hline 
        BadWeather & 6 & 6 & 0.21 & 8.00 & 33.33\% & \textbf{0.73} & \textbf{7.84} & \textbf{30.59\%} \\
        BringCoffee & 4 & 4 & \textbf{0.01} & \textbf{11.00} & \textbf{175.00\%} & 0.0 & -- & --\% \\
        BringCupsInterrupt & 10 & 14 & \textbf{0.1} & \textbf{16.60} & \textbf{66.00\%} & 0.0 & -- & --\% \\
        FindBackpack & 7 & 9 & 0.92 & 9.01 & 28.73\% & \textbf{0.94} & \textbf{8.24} & \textbf{17.78\%} \\
        GoToBanned & 14 & 20 & 0.0 & -- & --\% & 0.0 & -- & --\% \\
        HalloweenShopping & 12 & 18 & 0.0 & -- & --\% & 0.0 & -- & --\% \\
        MeetingTime & 5 & 5 & 0.0 & -- & --\% & 0.0 & -- & --\% \\
        PlacePlants & 11 & 15 & \textbf{0.47} & \textbf{15.26} & \textbf{38.68\%} & 0.0 & -- & --\% \\
        SearchHarder & 14 & 9 & 0.0 & -- & --\% & \textbf{0.09} & \textbf{16.44} & \textbf{17.46\%} \\
        WalkingBuddy & 5 & 11 & \textbf{0.12} & 9.17 & 83.33\% & 0.02 & \textbf{9.00} & \textbf{80.00\%} \\
        WhichPopcorn & 20 & 20 & \textbf{1.0} & \textbf{20.00} & \textbf{0.00\%} & 0.57 & \textbf{20.00} & \textbf{0.00\%} \\
    \end{tabular}
    \caption{The difference between the average steps taken to complete a task versus the optimal steps possible with perfect knowledge for \RoboRepair{} on Llama3.1-8B in both completion and chat mode. ``--'' means that the task did not complete. ``Perfect steps'' assumes no interruptions or errors.}
    \label{tab:llama3.1-8b-optimality}
\end{table*}

\begin{table*}[t]
    \centering
    \footnotesize
    \begin{tabular}{lrr|rrr|rrr}
        &&& \multicolumn{6}{|c}{DeepSeekCoder-V2-Lite}  \\
        \hline
        &&& \multicolumn{3}{|c}{Completion} & \multicolumn{3}{|c}{Chat}  \\
        \hline
                & Optimal & Ideal & & Average & & & Average & \\
        Problem & Steps & Steps & Pass@1 & Steps & \% Diff. & Pass@1 & Steps & \% Diff. \\
        \hline 
        BadWeather & 6 & 6 & 0.0 & -- & --\% & 0.0 & -- & --\% \\
        BringCoffee & 4 & 4 & 0.0 & -- & --\% & 0.0 & -- & --\% \\
        BringCupsInterrupt & 10 & 14 & 0.0 & -- & --\% & 0.0 & -- & --\% \\
        FindBackpack & 7 & 9 & 0.29 & \textbf{8.07} & \textbf{15.27\%} & \textbf{0.63} & 8.65 & 23.58\% \\
        GoToBanned & 14 & 20 & 0.0 & -- & --\% & 0.0 & -- & --\% \\
        HalloweenShopping & 12 & 18 & 0.0 & -- & --\% & 0.0 & -- & --\% \\
        MeetingTime & 5 & 5 & 0.0 & -- & --\% & 0.0 & -- & --\% \\
        PlacePlants & 11 & 15 & 0.0 & -- & --\% & 0.0 & -- & --\% \\
        SearchHarder & 14 & 9 & 0.21 & 16.05 & 14.63\% & \textbf{0.46} & \textbf{16.00} & \textbf{14.29\%} \\
        WalkingBuddy & 5 & 11 & \textbf{0.34} & \textbf{9.00} & \textbf{80.00\%} & 0.18 & \textbf{9.00} & \textbf{80.00\%} \\
        WhichPopcorn & 20 & 20 & \textbf{0.75} & \textbf{20.00} & \textbf{0.00\%} & 0.04 & \textbf{20.00} & \textbf{0.00\%} \\
    \end{tabular}
    \caption{The difference between the average steps taken to complete a task versus the optimal steps possible with perfect knowledge for \RoboRepair{} on DeepSeekCoder-V2-Lite in both completion and chat mode. ``--'' means that the task did not complete. ``Perfect steps'' assumes no interruptions or errors.}
    \label{tab:deepseek-optimality}
\end{table*}

\section{Optimality of Other Language Models Running \RoboRepair{}}
\label{appendix-results}

\cref{tab:llama3.1-8b-optimality,tab:deepseek-optimality} presents results with Llama 3.1 8B and DeepSeekCoder v2 Lite.

\section{Models and Compute Resources Used}
\label{models-and-compute}

 We evaluate Meta Llama 3.1 (8B and 70B)~\cite{dubey2024llama3herdmodels} and DeepSeek-Coder-V2 (16B)~\cite{deepseekai2024deepseekcoderv2breakingbarrierclosedsource}. Meta Llama 3.1 is under the Llama 3.1 Community License and DeepSeek-Coder-V2 is under the DeepSeek License Agreement. Both have an intended use of code generation and are allowed to be used in research settings. We run evaluation on an 8xH100 GPU server for 96 compute hours.
 
\end{document}